\documentclass[runningheads]{llncs}

\usepackage{graphicx}
\usepackage{subfiles}
\usepackage{todonotes}
\usepackage{comment}
\usepackage{amsmath,amssymb} % define this before the line numbering.
\usepackage{amsfonts}
\usepackage[misc,geometry]{ifsym}
\usepackage{color, soul}
\usepackage{color, colortbl}
\usepackage{multirow}
\usepackage{makecell}
\usepackage{xcolor}
\usepackage{array}
\usepackage{booktabs}
\usepackage{hyperref}
\usepackage{bm}
\definecolor{LightCyan}{rgb}{0.88,1,1}

\newcolumntype{L}{>{\centering\arraybackslash}m{0.1\textwidth}}
\newcolumntype{K}{>{\centering\arraybackslash}m{0.15\textwidth}}

\definecolor{bluegray}{RGB}{102, 153, 204}
\definecolor{steelblue}{RGB}{70, 130, 180}

\newif\ifdraft
\drafttrue

\definecolor{orange}{rgb}{1,0.5,0}
\definecolor{pink}{rgb}{0.98, 0.38, 0.5}
\definecolor{darkgreen}{rgb}{0.055, 0.490, 0.016} 

\ifdraft
 \usepackage[normalem]{ulem}
 \newcommand{\RS}[1]{{\color{red}{\bf RS: #1}}}
 
 \newcommand{\PMN}[1]{{\color{orange}{\bf PMN: #1}}}
 
 \newcommand{\STM}[1]{{\color{blue}{\bf STM: #1}}}

\else
 \newcommand{\sout}[1]{}
 \newcommand{\RS}[1]{{\color{red}{}}}
 
 \newcommand{\PMN}[1]{{\color{red}{}}}
 
 \newcommand{\STM}[1]{{\color{red}{}}}
 
  \newcommand{\ARGUMENT}[1]{{\color{gray}{}}}
 
\fi

\newcommand{\real}{\mathbb{R}}

\newcommand{\x}{\mathbf{x}}

\newcommand{\q}{\mathbf{q}}

\newcommand{\w}{\mathbf{w}}
\newcommand{\rbf}{\mathbf{r}}

\newcommand{\W}{\mathbf{W}}

\newcommand{\thetabf}{\bm{\theta}}

\newcommand{\ours}{\textbf{Ours}}

\begin{document}

\title{Targeted Visual Prompting for \\Medical Visual Question Answering}
\titlerunning{Targeted Visual Prompting for Medical Visual Question Answering}


%\author{Sergio Tascon-Morales \Letter, Pablo Márquez-Neila, Raphael Sznitman}
%index{Tascon-Morales, Sergio}
%index{Márquez-Neila, Pablo}
%index{Sznitman, Raphael}

\author{Sergio Tascon-Morales \Letter, Pablo Márquez-Neila, Raphael Sznitman}
% index{Tascon-Morales, Sergio}
% index{Márquez-Neila, Pablo}
% index{Sznitman, Raphael}

\authorrunning{Tascon-Morales et al.}

\institute{University of Bern, Bern, Switzerland\\ \email{\{sergio.tasconmorales, pablo.marquez, raphael.sznitman\}@unibe.ch}}

\maketitle          

\begin{abstract}

With growing interest in recent years, medical visual question answering (Med-VQA) has rapidly evolved, with multimodal large language models (MLLMs) emerging as an alternative to classical model architectures. Specifically, their ability to add visual information to the input of pre-trained LLMs brings new capabilities for image interpretation. However, simple visual errors cast doubt on the actual visual understanding abilities of these models. To address this, region-based questions have been proposed as a means to assess and enhance actual visual understanding through compositional evaluation. To combine these two perspectives, this paper introduces targeted visual prompting to equip MLLMs with region-based questioning capabilities. By presenting the model with both the isolated region and the region in its context in a customized visual prompt, we show the effectiveness of our method across multiple datasets while comparing it to several baseline models. Our code and data are available at \url{https://github.com/sergiotasconmorales/locvqallm}.


\end{abstract}
\section{Introduction}
\label{sec:intro}

% introductory paragraph
Visual Question Answering (VQA) is centered on developing models capable of answering questions about specific images~\cite{antol2015vqa}. This task is particularly challenging within the medical domain due to factors such as a scarcity of annotated data~\cite{Nguyen19,liu2019effective}, the wide variety of imaging modalities and anatomical regions~\cite{gupta2021hierarchical}, as well as the unique characteristics of medical images and terminology, all of which necessitate specialized expertise~\cite{liu2019effective,zhan2020medical}. Furthermore, approaches that leverage the detection of natural objects, which have significantly improved performance in the analysis of natural images~\cite{anderson2018bottom}, are less straightforward when applied to medical imagery~\cite{gupta2021hierarchical}.

Historically, models for Medical VQA (Med-VQA) treated visual and textual information independently, later merging these features through various fusion techniques. This composite data would then be input into a classifier to determine the most probable answer. However, recent developments in transformer-based models~\cite{vaswani2017attention}, including advancements in Large Language Models (LLMs), have led to a notable shift in VQA strategies. These advancements have paved the way for the adoption of multimodal LLMs (MLLMs) that integrate both visual and textual data more seamlessly, a trend that is emerging in both general~\cite{yin2023survey,tong2024eyes,zhang2024mm} and Med-VQA~\cite{seenivasan2023surgicalgpt,zhang2023pmc} applications.

\begin{figure}[!t]
\begin{center}
\includegraphics[width=0.95\textwidth]{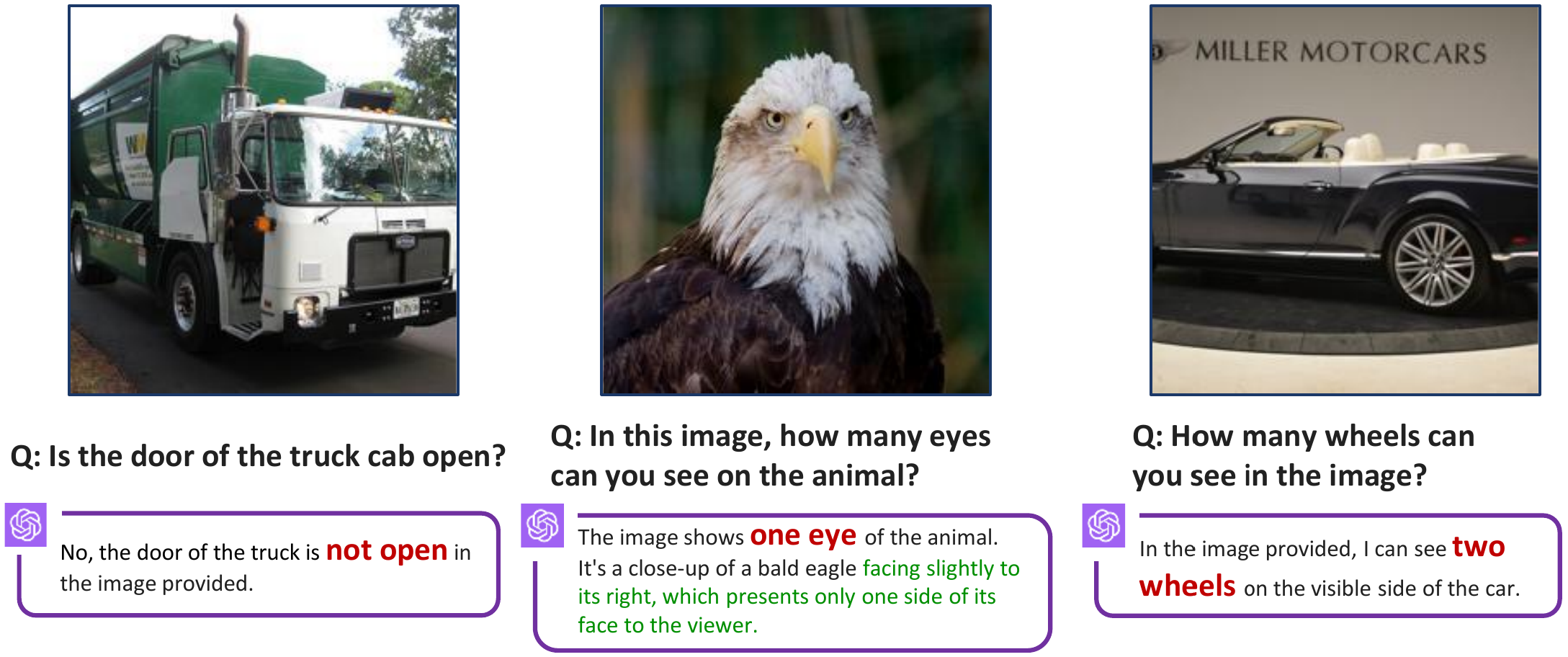}
\caption{Examples of visual understanding failures using GPT-4V for the VQA task (Examples taken from~\cite{tong2024eyes}).}
\label{fig:examples_gpt_4v}
\end{center}
\end{figure}

% About models not actually understanding images
Despite the remarkable adoption of MLLMs, recent research has raised concerns about the quality of their visual capabilities (Fig.~\ref{fig:examples_gpt_4v}). This issue primarily arises from the pre-training process of the visual component, which typically relies on models like CLIP~\cite{radford2021learning}. Surprisingly, MLLMs can perceive certain visually distinct images as similar, a phenomenon that human observers readily recognize as a visual error~\cite{tong2024eyes}. These visual understanding failures were also observed in VQA models before the widespread adoption of MLLMs~\cite{goyal2017making,hudson2019gqa,ribeiro2019red,selvaraju2020squinting}.

To detect such failures and enhance explainability in the visual component of Med-VQA, the work in ~\cite{tascon2023localized} proposes a novel approach using the formulation of \textit{localized questions}~\cite{tascon2023localized}. These questions allow fine-grained probing of images by focusing on user-defined regions rather than the entire image and facilitate a \textit{compositional evaluation}. To enable such localized questions, the region to query is encoded and directly integrated into the attention mechanism of the model. Other proposed strategies include providing the model with a restricted region of the image~\cite{tascon2022consistency} or relying on the language component of the VQA model to interpret region coordinates directly included in the question~\cite{vu2020question}. Yet, due to their design focused on traditional architectures, these methods fail to benefit MLLMs in Med-VQA. Other traditional~\cite{mani2020point} and MLLM-based methods~\cite{chen2023position,zhang2023gpt4roi} rely on object detectors, limiting their applicability to medical images. 

%For example, consider the car image on the right in Fig.~\ref{fig:examples_gpt_4v}. Instead of asking a generic question like "How many wheels can you see?" about the entire image, we can pose the same question specifically for regions that do not contain any wheels. 

To overcome these challenges and enable localized questions in MLLMs in Med-VQA, we introduce {\it Targeted Visual Prompting}. By carefully designing a prompt that provides both global and local visual tokens relative to the region of interest defined by the user, our method allows the full advantage of the MLLM to enhance the performance of the VQA model. To validate the effectiveness of our method, we conduct exhaustive experiments across multiple datasets. Our results demonstrate clear performance benefits compared to previously proposed methods, all achieved without introducing additional parameters to the model. %By bridging the gap between MLLMs and domain-specific requirements, Targeted Visual Prompting represents a significant step forward in Med-VQA research.

\section{Method}
\label{sec:method}

A VQA model with parameters $\thetabf$ generates an answer $\hat{a}$ when given an input image $\x \in \real^{H\times W\times C}$ and a related question represented as a sequence of words, $\q$. %Here, $H$, $W$, and $C$ denote the height, width, and channels of the image, respectively. 
In its most general form, this process can be described as a function $\Psi_{\text{VQA}}$, parameterized by $\thetabf$, that is applied on the image-question pair,
\begin{equation}
    \hat{a} = \Psi_{\text{VQA}} (\x, \q; \thetabf).
\end{equation}
In practice, this model's output has traditionally been a distribution over a set of $N$ candidate answers $\{a_1, a_2, ..., a_N\}$ set beforehand. 

In this work, however, we choose the answer of the VQA to be generated by an LLM in an auto-regressive manner until the end-of-sentence (EOS) token is produced. To make the LLM multimodal, we adopt the widely used approach of projecting visual embeddings onto the input space of the LLM~\cite{liu2023visual,tsimpoukelli2021multimodal,wang2023r2gengpt} and 
express this as, 
\begin{equation}
    \hat{a} = \Psi_{\text{LLM}} (\Psi_{\text{Vis}} (\x, \thetabf_{\text{Vis}})\W^\text{proj}, \q; \thetabf_{\text{LLM}}),
\end{equation}
\noindent
where $\Psi_{\text{Vis}}$ refers to the visual encoder with parameters $\thetabf_{\text{Vis}}$, and $\W^\text{proj}$ denotes the learnable parameters of the projection layer. Although not explicitly formalized, it is implied that the answer is generated in an auto-regressive fashion, meaning that the next word in the answer depends on the previously predicted words.

To expand the model's capability to handle localized questions, we propose here a dedicated targeted visual prompt that allows two perspectives of the image to be encoded: one containing only the region of the image and the other containing the region in context. 

\begin{figure}[!t]
\begin{center}
\includegraphics[width=\textwidth]{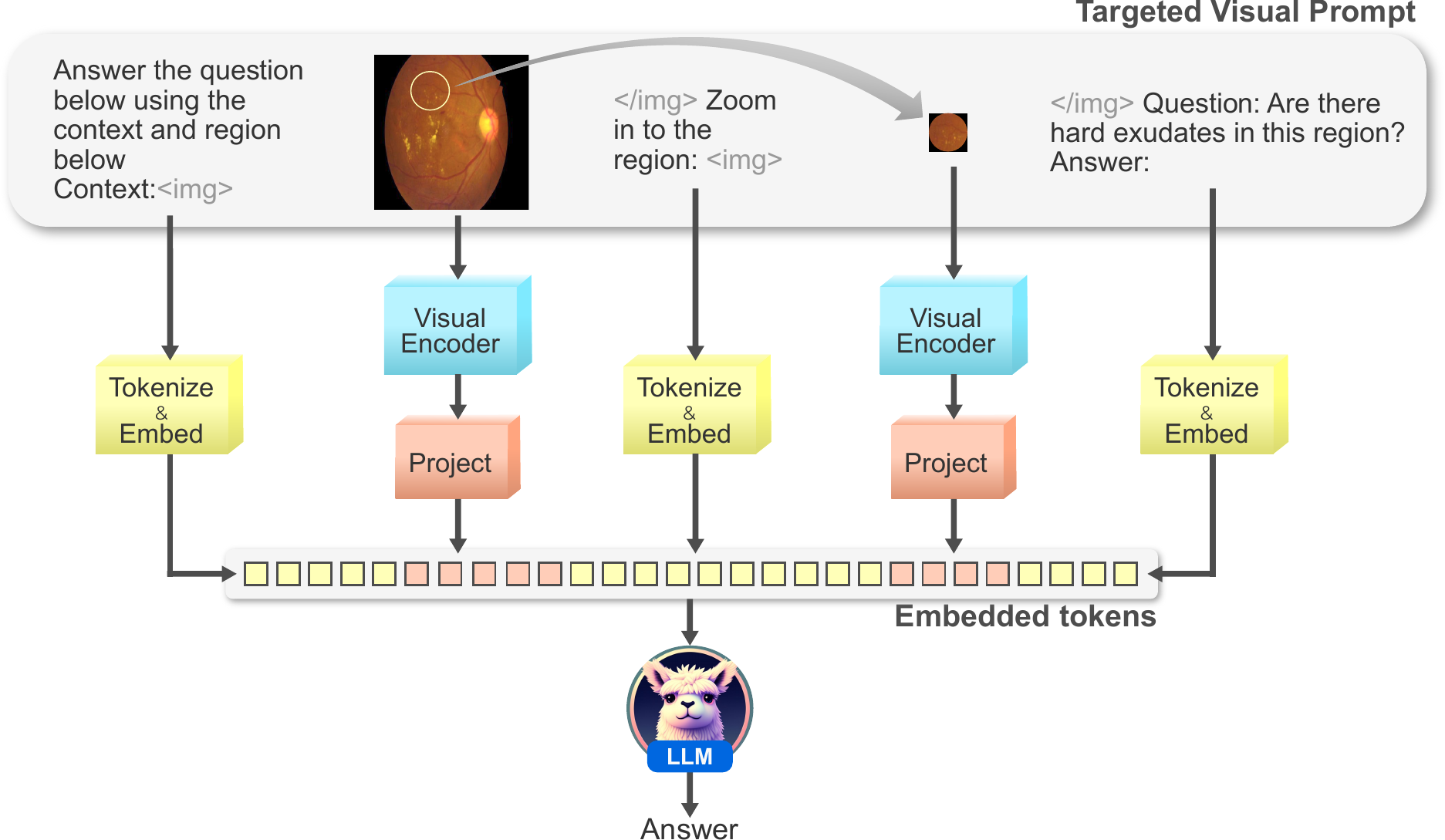}
\caption{Our customized targeted visual prompt is created by providing the model with the region in context, as well as an isolated version of the region. Visual tokens are projected to the input space of the LLM and concatenated with the instruction tokens.}
\label{fig:method}
\end{center}
\end{figure}

The targeted visual prompt consists of five components: (1) comprises model instruction, denoted as $\w_{\text{instr}}$; (2) the visual context represented by the image with the region drawn on it, $\x_r$; (3) $\w_\text{det}$ contains a textual prefix for the region; (4) the cropped region $\rbf$; and (5) $\w_q$ includes the question $\q$. Text-containing parts of the prompt undergo tokenization and embedding, while the visual components are processed by a visual encoder and then projected into the input space of the LLM. Subsequently, the results are concatenated and processed by the LLM, resulting in the generation of an answer. To handle global questions, the entire image is assigned to $\rbf$. We illustrate our model in Fig.~\ref{fig:method} and summarize the computation of the answer as,
\begin{equation}
    \hat{a} = \Psi_{\text{LLM}} (\w_{\text{instr}}, \Psi_{\text{Vis}} (\x_r, \thetabf_{\text{Vis}})\W_{\x_r}^\text{proj}, \w_\text{det}, \Psi_{\text{Vis}} (\rbf, \thetabf_{\text{Vis}})\W_{\rbf}^\text{proj}, \w_q; \thetabf_{\text{LLM}}).
\end{equation}

To handle questions about the entire image, both $\x_r$ and $\rbf$ correspond to the original image.

\textbf{Training. } As in~\cite{wang2023r2gengpt}, our model is trained using the original auto-regressive training loss of the LLM. The loss function is the standard negative log-likelihood accumulated over all time steps for predicting the correct next token. For a ground truth answer of length $T$, this loss is expressed as,
\begin{equation}
    \mathcal{L}(\thetabf) = - \sum_{t=1}^T \log p_\theta(a^t | \x, \w, a^{1:t-1}; \thetabf),
\end{equation}
\noindent
where $\x$ and $\w$ denote the visual and textual elements, respectively, and $\mathbf{a} = \{a_1, a_2, ..., a_T\}$ is the ground truth answer.

\section{Experiments and results}
\label{sec:experiments_and_results}

\textbf{Datasets: } To evaluate our method, we make use of several publically available datasets~\cite{tascon2023localized}: (1) DME-VQA: contains questions on diabetic macular edema (DME) risk grade and about the presence of biomarkers in the entire image or specific regions. (2) RIS-VQA: contains images from the DaVinci robot during gastrointestinal surgery and questions related to surgical instruments. (3) INSEGCAT-VQA: contains frames from cataract surgery videos and questions about instruments used in this type of surgery. A summary of these is shown in Table~\ref{tab:datasets}. For all datasets, we use the same partitioning as in~\cite{tascon2023localized}.

\begin{table}[ht]
\begin{center}
\begin{tabular}{lp{0.5cm}lp{0.5cm}cp{0.5cm}c}
\toprule
Dataset                       && Modality         && \# images && \# QA-pairs \\ \midrule
DME-VQA                        && Fundus && 679 && 13470 \\
RIS-VQA                        && Gastrointestinal && 2978 && 32562 \\
INSEGCAT-VQA                   && Cataract surgery && 4647 && 39008 \\
\bottomrule
\end{tabular}
\end{center}
\caption{Dataset parameters.}
\label{tab:datasets}
\end{table}

\begin{figure}[t]
\begin{center}
\includegraphics[width=\textwidth]{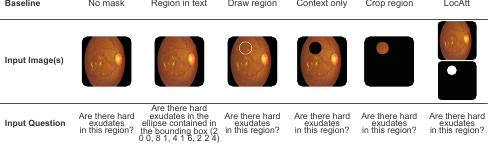}
\caption{Example input images and questions for evaluated baselines. In the baseline ``Region in text," the digits are separated to provide a fair scenario to the LLM.}
\label{fig:baselines}
\end{center}
\end{figure}

\textbf{Baselines:} We benchmark our method against multiple baselines, which are exemplified in Fig.~\ref{fig:baselines}. In \textbf{No mask}, the model receives no information about the location of the region; in \textbf{Region in text}, the region is specified in the question; in \textbf{Draw region}, the region is marked on top of the image. In \textbf{Context only}, the model only sees the context, but not the contents of the region; in \textbf{Crop region}, the model receives no context; finally, in \textbf{LocAtt}~\cite{tascon2023localized}, the model has access to the image, as well as a binary image representing the region. For these baselines, the visual prompt given to the model is: \textit{``Answer the question below using the context below Context: $<$Img$><$Image$><$/Img$>$Question:$<$Question$>$Answer:"}

\subsubsection{Implementation details:} We use R2GenGPT~\cite{wang2023r2gengpt} as base MLLM, adapting it from the task of radiology report generation to VQA. We use a pre-trained Swin Transformer~\cite{liu2021swin} as visual encoder and Llama 2 7B~\cite{touvron2023llama} as LLM, initialized with its official weights. Different from to R2GenGPT, we finetune all modules, including the LLM, end-to-end. We train all our models for 15 epochs, with a batch size of 8 and learning rate of 1e-4, with the AdamW optimizer and a cosine annealing scheduler with minimum learning rate 1e-6. For the text generation, we use a repetition penalty of 2.0 and a length penalty of -1.0. Our implementation uses PyTorch 2.0.1 and two Nvidia A100 cards with 80 GB of memory each.

\subsection{Results} 
\label{sec:results}

Table~\ref{tab:results} summarizes our results on the DME-VQA, RIS-VQA, and INSEGCAT-VQA datasets. The accuracy and F1 score are reported for all datasets. Notably, our method consistently outperforms all evaluated baselines across all datasets, underscoring the efficacy of targeted visual prompting in enhancing MLLMs with localized question capabilities. 

\begin{table}[!t]
\begin{center}
\begin{tabular}{lp{0.5cm}lp{0.5cm}cp{0.5cm}cp{0.5cm}}
\toprule
Dataset                       && Method         && Accuracy (\%) && F1 score (\%) \\ \midrule
\multirow{7}{*}{DME-VQA}      && No Mask    && 57.32 &&    57.32   \\ 
                                && Region in Text~\cite{vu2020question} &&  62.12 &&  63.59 \\ 
                              && Crop Region~\cite{tascon2022consistency}    &&  86.52   &&  87.26   \\ 
                              %&& Adapted LocAtt~\cite{tascon2023localized}$^\dag$  && 86.30 \\
                              && Draw Region && 86.86 && 86.85 \\ 
                              && Context Only && 88.07  && 88.45 \\ 
                              && \ours           && \textbf{90.30} && \textbf{90.22} \\ 
                             && \textcolor{gray}{LocAtt~\cite{tascon2023localized}$^*$}  && \textcolor{gray}{84.2} && \textcolor{gray}{85.79}  \\
\midrule
\multirow{7}{*}{RIS-VQA}      && No Mask    && 50.00  &&  50.00   \\ 
                                && Region in Text~\cite{vu2020question} &&  64.81 &&  65.39\\ 
                              && Crop Region~\cite{tascon2022consistency}    &&  85.50  &&   85.64   \\ 
                              %&& Adapted LocAtt~\cite{tascon2023localized}$^\dag$  && 86.30 \\
                              && Draw Region && 91.30  && 91.43 \\ 
                              && Context Only && 91.77 && 91.81  \\ 
                              && \ours           && \textbf{92.60} && \textbf{92.54} \\ 
                             && \textcolor{gray}{LocAtt~\cite{tascon2023localized}$^*$}  && \textcolor{gray}{82.73} &&  \textcolor{gray}{86.15}\\
                              \midrule
\multirow{7}{*}{INSEGCAT-VQA} && No Mask    &&   50.00  &&  50.00  \\  
                              && Region in Text~\cite{vu2020question} &&  73.51  && 74.55 \\ 
                              && Crop Region~\cite{tascon2022consistency}    &&  90.91  &&  90.93  \\ 
                              %&& Adapted LocAtt~\cite{tascon2023localized}$^\dag$  && 89.89 \\
                              && Draw Region && 95.44 && 95.43\\  
                            && Context Only && 95.19 &&  95.17 \\ 
                              && \ours           &&  \textbf{95.51}    &&  \textbf{95.47} \\
                              && \textcolor{gray}{LocAtt~\cite{tascon2023localized}$^*$}  && \textcolor{gray}{88.13} && \textcolor{gray}{90.14}\\                              
                              \bottomrule
\end{tabular}
\end{center}
\caption{Accuracy and F1 score comparison to SOTA approaches on the DME-VQA, RIS-VQA and INSEGCAT-VQA datasets. For the DME-VQA dataset only localized questions are considered. $^*$This result corresponds to a different architecture, but we include it for completeness. }
\label{tab:results}
\end{table}

%In general, it is observed that our method is most effective in questions for which both context and the contents of the region offer meaningful complementary information. This occurs in the DME-VQA dataset, where the reduced size of the biomarkers makes the contents of the region crucial, but at the same time the presence of biomarkers in other parts of the image increase the chances of having regions with biomarkers. 

In the case of the DME-VQA and RIS-VQA datasets, we observe that the performance of \textit{context only} surpasses that of \textit{crop region}. At first glance, this suggests that the context holds more relevance than the specific contents of the region. However, this behavior is likely influenced by spurious correlations between region sizes/locations, and the corresponding answers. For instance, in DME-VQA, images with a high amount of biomarkers often feature smaller regions associated with negative answers. Another reason for this behavior is that in many cases the context provides more evidence, in terms of pixel count, to answer the question, as compared to the region. For instance, in RIS-VQA, the tool can often be determined from its body without considering the tip. 

Notably, the \textit{region in text} baseline exhibits poor performance. Given the use of a powerful LLM in the pipeline, higher performance might be expected. Different variations were explored for this baseline, including not separating the coordinate digits or replacing coordinate digits with words, but performance did not improve. We hypothesize that the model fails to correctly map location information from the text to the image, which can be at least partly attributed to using a ViT to embed the image.

\begin{figure}[!t]
\begin{center}
\includegraphics[width=\textwidth]{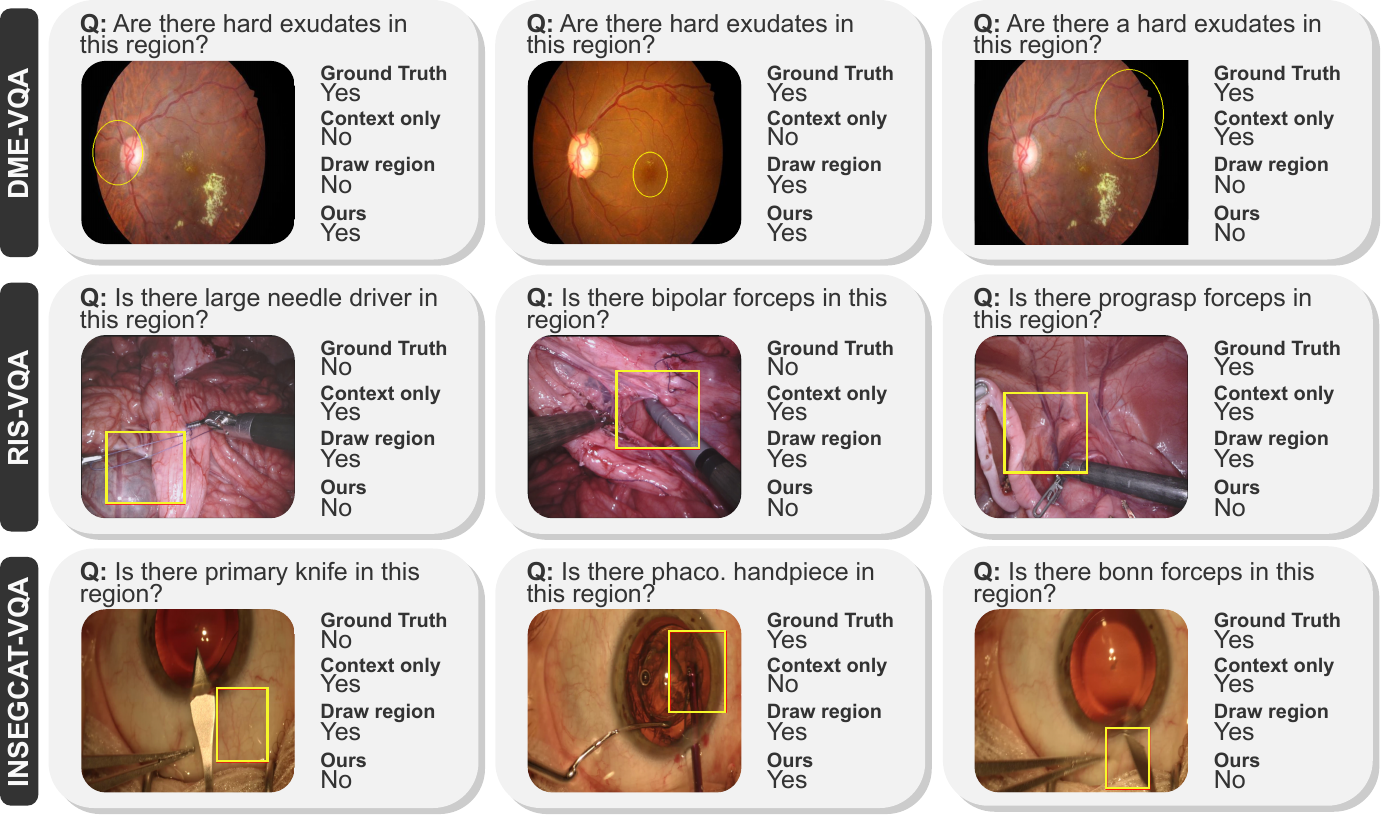}
\caption{Qualitative examples on the DME-VQA (first row), RIS-VQA (second row), and INSEGCAT-VQA (third row) datasets.}
\label{fig:examples}
\end{center}
\end{figure}

We provide qualitative example results in Fig.~\ref{fig:examples}. The first column exemplifies cases where our method demonstrates robustness to subtle evidence (small biomarkers), correlations (surgical suture is usually close to the needle driver), and borderline cases (evidence close to the region border). The second column highlights the weaknesses of \textit{context only} when the context fails to provide enough evidence for the answer. Finally, the third column shows errors made by our model in tricky cases (subtle or ambiguous evidence in the region).

\begin{figure}[!t]
\begin{center}
\includegraphics[width=0.99\textwidth]{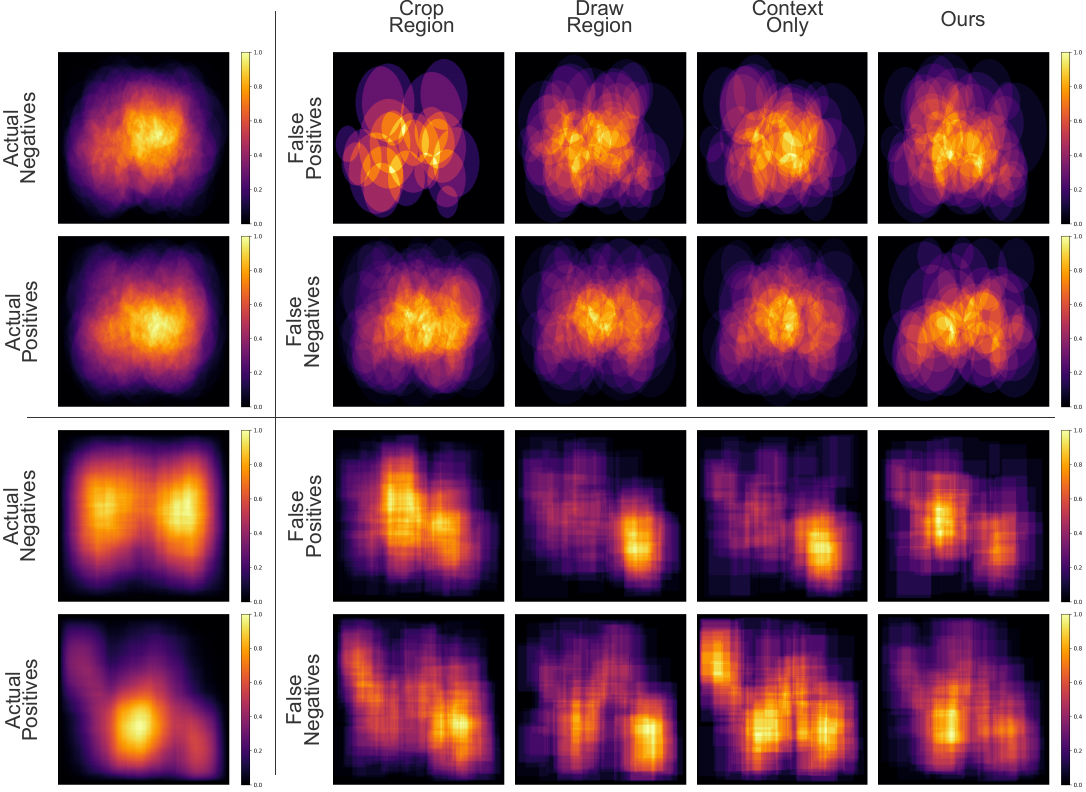}
\caption{Error analysis by region location for the four strongest baselines. The maps are obtained by adding binary masks representing the regions for all QA pairs in each category and then normalizing. \textbf{Top:} DME-VQA dataset. \textbf{Bottom:} INSEGCAT-VQA dataset.}
\label{fig:errors_by_location}
\end{center}
\end{figure}

Fig.~\ref{fig:errors_by_location} shows error maps by region location for the DME-VQA and INSEGCAT-VQA datasets and for the four strongest baselines. On the left side of the plot, the locations of actual positives and negatives are illustrated. For the INSEGCAT-VQA dataset, this visualization reveals a location bias that other baselines without access to the region or the context may be exploiting. Due to the nature of the images (cataract surgery) and questions, regions with positive answers tend to cluster in a specific area. This, coupled with the dissimilarity of objects mentioned in the questions, explains why a baseline like \textit{crop region} achieves relatively high performance on this dataset compared to the other two datasets (see Table~\ref{tab:results}). Similarly, in the case of DME-VQA, it becomes evident that the lack of context in \textit{crop region} results in lower sensitivity, highlighting the significance of context even when the isolated region should theoretically provide sufficient evidence. Fig.~\ref{fig:errors_by_location} also demonstrates that \textit{draw region} and \textit{context only} exhibit marked clusters of false positives and negatives in INSEGCAT-VQA, potentially indicating the utilization of location biases. In contrast, our method produces a more evenly distributed location for both types of errors.

%\begin{figure}[!t]
%\begin{center}
%\includegraphics[width=\textwidth]{images/results_by_region_size.pdf}
%\caption{Results of our proposed method by region size for all datasets. Horizontal axes show the region size as a percentage of the image size. Numbers within each bar indicate the amount of QA pairs that fall into each bin. \STM{Not sure if this image is necessary or useful. The goal was to show that the model is robust to region size.}}
%\label{fig:results_by_region_size}
%\end{center}
%\end{figure}

\section{Conclusions}
\label{sec:conc}

In this work, we introduced a novel approach to enable localized questions in multimodal LLMs for the tasks of VQA. Our proposed approach involves the utilization of targeted visual prompting, granting the model access not only to the region and its context within the image but also to an isolated version of the region. Doing so allows two perspectives to be encoded in the prompt and more fine-grained information to be leveraged. Our approach demonstrates enhanced performance across all evaluated datasets compared to various baselines. Future works include extending the methodology to accommodate multiple images and enabling the use of comparison questions.

\textbf{Prospect of application:} This approach aims to be useful for medical assistants/chatbots that can help doctors assess specific parts of an image that look suspicious. By providing a second opinion, it can improve the accuracy of diagnoses. Additionally, this technology could help medical students learn and reinforce medical concepts by enabling a more modular analysis of medical images.

\begin{credits}
\subsubsection{\ackname} This work was partially funded by the Swiss National Science Foundation through grant 191983.

\subsubsection{\discintname} 
The authors have no competing interests to declare that are relevant to the content of this article.
\end{credits}

%\clearpage 
\bibliographystyle{splncs04}
\bibliography{mybibliography}

\end{document}

\title{Supplementary Material\\Targeted Visual Prompting for Medical Visual Question Answering}
\author{Anonymous}
\authorrunning{F. Author et al.}
\institute{Anonymous Organization\\ \email{email}}
\maketitle              % typeset the header of the contribution
%

%\subfile{sections/00_Abstract}

\begin{table}[!h]
\begin{center}
\begin{tabular}{llp{0.1cm}lp{0.1cm}lp{0.1cm}l}
\toprule
\multicolumn{1}{c}{\multirow{2}{*}{Method}} & \multicolumn{7}{c}{Accuracy (\%)}                                                                                                                                               \\ \cmidrule{2-8} 
\multicolumn{1}{c}{} & \multicolumn{1}{c}{Overall}      && \multicolumn{1}{c}{Grade}        && \multicolumn{1}{c}{Whole}        && \multicolumn{1}{c}{Macula}      \\ \midrule 
No Mask & \multicolumn{1}{c}{60.50} && \multicolumn{1}{c}{81.13} && \multicolumn{1}{c}{76.42} && \multicolumn{1}{c}{85.85}                \\ 
Region in Text & \multicolumn{1}{c}{ 64.75} && \multicolumn{1}{c}{79.25} && \multicolumn{1}{c}{83.96} && \multicolumn{1}{c}{ 82.08}        \\ 
Crop Region  & \multicolumn{1}{c}{86.05} && \multicolumn{1}{c}{80.19} && \multicolumn{1}{c}{83.96} && \multicolumn{1}{c}{84.91}                \\
Draw Region  & \multicolumn{1}{c}{ 86.18} && \multicolumn{1}{c}{79.25} && \multicolumn{1}{c}{83.02} && \multicolumn{1}{c}{83.02}              \\
Context Only  & \multicolumn{1}{c}{ 82.61} && \multicolumn{1}{c}{76.42} && \multicolumn{1}{c}{87.74} && \multicolumn{1}{c}{90.57}              \\
\ours                                          & \multicolumn{1}{c}{89.29} && \multicolumn{1}{c}{79.25} && \multicolumn{1}{c}{83.96} && \multicolumn{1}{c}{84.91}              \\ \bottomrule
\end{tabular}
\end{center}
\caption{Accuracy for the DME-VQA dataset by question type.}
\label{tab:results_dme}
\end{table}

\begin{figure}[!h]
\begin{center}
\includegraphics[width=0.95\textwidth]{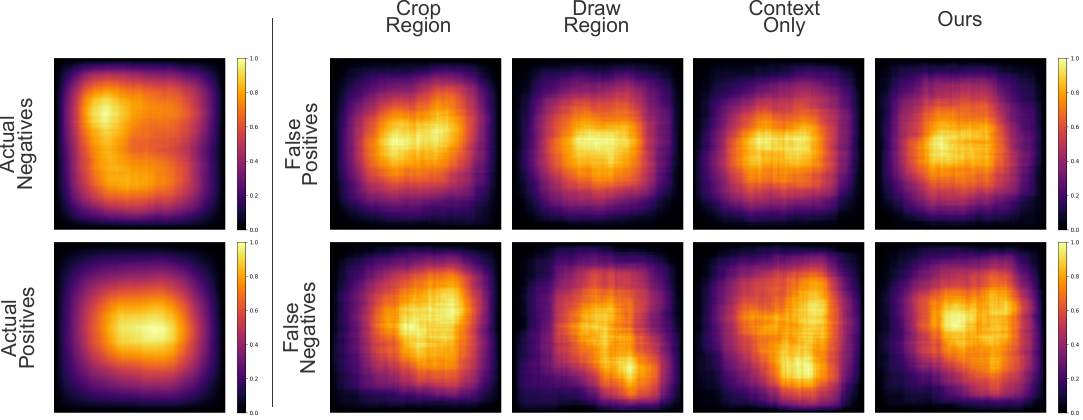}
\caption{Error analysis by region location for the four strongest baselines for the RIS-VQA dataset. The maps are obtained by adding binary masks representing the regions for all QA pairs in each category and then normalizing.}
\label{fig:supplementary_errors_by_location}
\end{center}
\end{figure}

\begin{figure}[!h]
\begin{center}
\includegraphics[width=\textwidth]{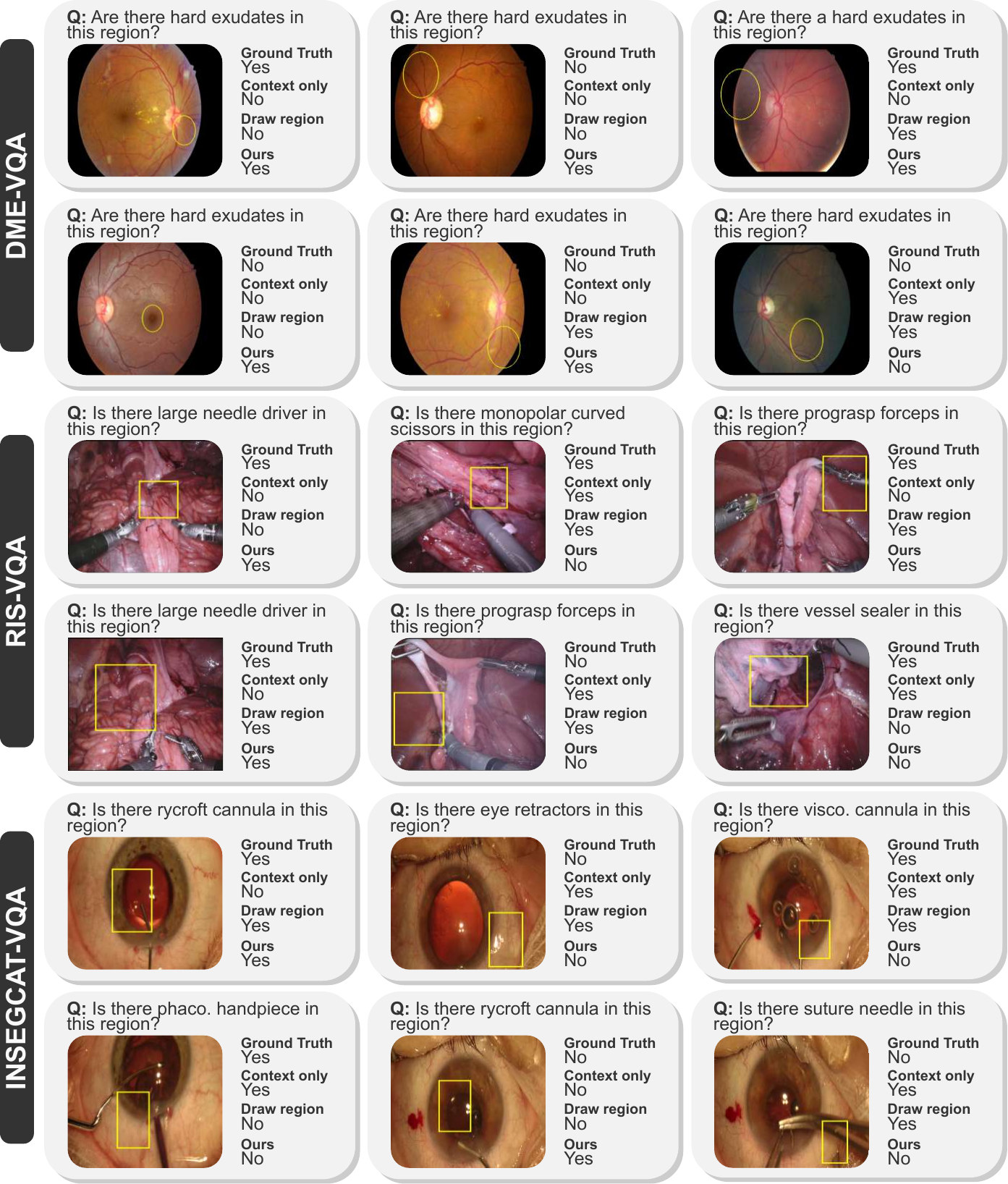}
\caption{Additional examples for DME-VQA (rows 1 and 2), RIS-VQA (rows 3 and 4) and Insegcat-vQA (rows 5 and 6).}
\label{fig:examples_supplementary_dme}
\end{center}
\end{figure}


%